\documentclass[]{onethree}

\usepackage[toc,page,header]{appendix}
\usepackage{minitoc}
\usepackage{amsmath,amssymb}
\usepackage{bm}
\usepackage{wrapfig}
\usepackage{enumitem}

\usepackage[numbers]{natbib}
\usepackage{amsmath,amssymb,amsfonts}
\usepackage{algorithmic,algorithm}
\usepackage{hyperref}
\usepackage{microtype}
\usepackage{booktabs}
\usepackage{multicol}
\usepackage{multirow}
\usepackage{array}
\usepackage{graphicx}
\usepackage{url}

\title{DeBoneDiT: Depth-Driven Conditional Bridge Diffusion Transformers for Bone Suppression in Chest X-Ray Images}

\author[1,2,*]{Yifei Sun}
\author[3,*]{Fenglei Fan}
\author[2]{Junhao Jia}
\author[4]{Wenming Deng}
\author[1]{Hongxia Xu}
\author[5,\dagger]{Changmiao Wang}
\author[2,\dagger]{Ruiquan Ge}

\affiliation[1]{Zhejiang University}
\affiliation[2]{Hangzhou Dianzi University}
\affiliation[3]{City University of Hong Kong}
\affiliation[4]{National Cancer Center/National Clinical Research Center for Cancer/Cancer Hospital \& Shenzhen Hospital, Chinese Academy of Medical Sciences and Peking Union Medical College}
\affiliation[5]{Shenzhen Research Institute of Big Data}

\contribution[*]{Equal contribution}
\contribution[\dagger]{Corresponding authors}

\abstract{
Chest X-Ray (CXR) is a primary modality for diagnosing pulmonary diseases, yet the overlap of bony structures often obscures critical radiographic findings, increasing the risk of diagnostic uncertainty. While Dual-Energy Subtraction (DES) is the clinical benchmark for bone suppression, its widespread adoption is primarily hindered by specialized hardware requirements and increased radiation exposure. Recently, diffusion-based generative models have shown promise for this task, yet they are still constrained by reconstruction-biased latent compression, prior-induced domain gap, and a scarcity of high-quality paired datasets. To address these challenges, we propose DeBoneDiT, a depth-driven conditional bridge diffusion Transformer architecture for efficient high-resolution bone suppression. We first introduce Depth Auto-Encoder (DAE), a structure-aware vision tokenizer for latent compression. By incorporating a multi-level depth feature loss derived from the pretrained Depth Anything V2 encoder, DAE preserves both visual fidelity and structural consistency. Second, we formulate bone suppression as a source-conditioned Brownian bridge diffusion process, leveraging CXRs as both the diffusion prior and structural guidance to narrow the source-to-target domain gap and maintain anatomical plausibility. Furthermore, we have constructed and released SZCH-X-Rays, the largest publicly available high-quality paired dataset for bone suppression to date, comprising 741 pairs of CXR and DES soft tissue images. Extensive experiments and downstream evaluation demonstrate that DeBoneDiT achieves superior bone suppression performance with reduced computational overhead.
\vspace{-0.1cm}
}

\checkdata[Email]{
  \href{mailto:diaoquesang@gmail.com}{diaoquesang@gmail.com}\hspace{2em}
}

\begin{document}
\maketitle
\vspace{-1.1cm}


\newpage

\tableofcontents
\newpage


\section{Introduction}

Lung diseases remain a leading cause of morbidity and mortality worldwide, posing substantial challenges to public health systems and healthcare resource allocation~\cite{oh2026global}. In the clinical diagnosis and screening of lung diseases, Chest X-Ray (CXR) serves as a primary imaging modality due to its accessibility, low cost, and minimal radiation exposure. It enables rapid detection of key radiographic findings such as pulmonary inflammation, nodules, and masses, providing crucial imaging evidence for early diagnosis. However, overlapping bony structures often obscure pathological details in critical regions, significantly increasing the risk of overlooking subtle early-stage lesions~\cite{gefter2023reducing}. In response to this challenge, bone suppression has attracted increasing attention as an effective strategy for enhancing the diagnostic utility of CXR. Clinical studies have demonstrated that bone-suppressed CXR images can significantly improve radiologists' sensitivity in detecting pulmonary lesions~\cite{kim2023development}. Moreover, bone-suppressed images provide cleaner and more informative inputs for Computer-Aided Detection (CAD) systems, thereby facilitating more accurate automated analysis and clinical decision-making~\cite{sun2025bs}.

Dual-Energy Subtraction (DES) imaging~\cite{vock2009dual} is the clinical benchmark for CXR bone suppression. By leveraging the differential X-ray attenuation of bone and soft tissue across two energy levels, it separates bone and soft tissue components to generate bone-suppressed images. Nevertheless, the widespread adoption of DES remains limited due to its reliance on specialized instrumentation, increased radiation exposure, and susceptibility to motion-induced artifacts. These limitations have spurred extensive research into alternative bone suppression approaches, with the goal of achieving effective bone removal while preserving diagnostically relevant soft tissue details.

Early bone suppression methods primarily relied on shallow pixel-level learning and statistical anatomical modeling to isolate and remove bony structures. Specifically, Suzuki et al.~\cite{suzuki2006image} employed a Massive Training Artificial Neural Network (MTANN) to generate bone residuals for subsequent subtraction, while Juh{\'a}sz et al.~\cite{juhasz2010segmentation} used an Active Shape Model (ASM) to segment anatomical structures. However, the clinical applicability of these methods remains limited, as they either lack high-level semantic understanding of bony anatomy or rely on labor-intensive boundary annotations.

Conventional deep learning approaches, including Auto-Encoder (AE)-based models~\cite{gusarev2017deep}, U-Net variants~\cite{schiller2025xu}, and Generative Adversarial Network (GAN)-driven frameworks~\cite{yang2025improving, singh2024mda, rani2022multi, zhou2018generation}, have since emerged to overcome the limitations of early methods through automated feature representation learning. Gusarev et al.~\cite{gusarev2017deep} treated bones as noise and used AEs to generate soft tissue images. For U-Net-based architectures, Schiller et al.~\cite{schiller2025xu} developed xU-NetFullSharp, incorporating multi-level and multi-directional feature connections. Among GAN-driven frameworks, Zhou et al.~\cite{zhou2018generation} proposed a Multi-Scale Conditional Adversarial Network (MCA-Net) to generate virtual DES soft tissue images; Rani et al.~\cite{rani2022multi} developed a Pix2Pix-based pipeline for bone suppression, aiming to support multi-modal downstream tasks; Singh et al.~\cite{singh2024mda} further enhanced GAN performance with a Multi-Scale and Dual Attention GAN (MDA-GAN); and Yang et al.~\cite{yang2025improving} advanced GAN-based methods by exploiting both domain-level and instance-level information. Despite these advances, many methods still face challenges in balancing effective bone suppression with the preservation of fine pulmonary structures, often resulting in over-smoothing, texture loss, or incomplete bone removal.

\begin{figure}[!t]
    \centering
    \includegraphics[width=0.8\linewidth]{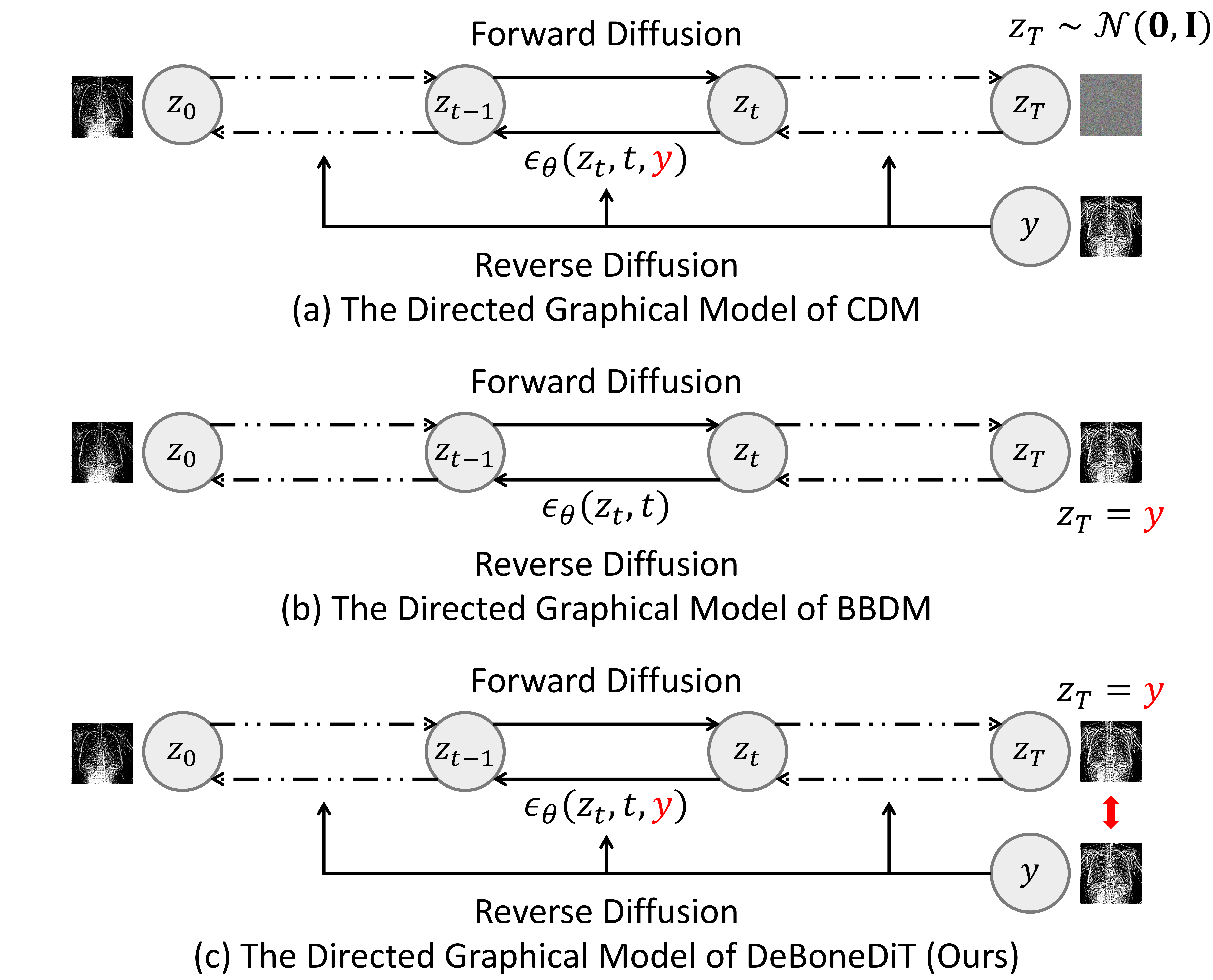}
    \caption{Comparison of directed graphical models of Conditional Diffusion Model (CDM)~\cite{ho2020denoising}, Brownian Bridge Diffusion Model (BBDM)~\cite{li2023bbdm} and DeBoneDiT (Ours). ($z_0$: latent tissue; $y$: latent CXR)}
    \label{paradigms}
\end{figure}

\begin{figure}[!t]
    \centering
    \includegraphics[width=0.8\linewidth]{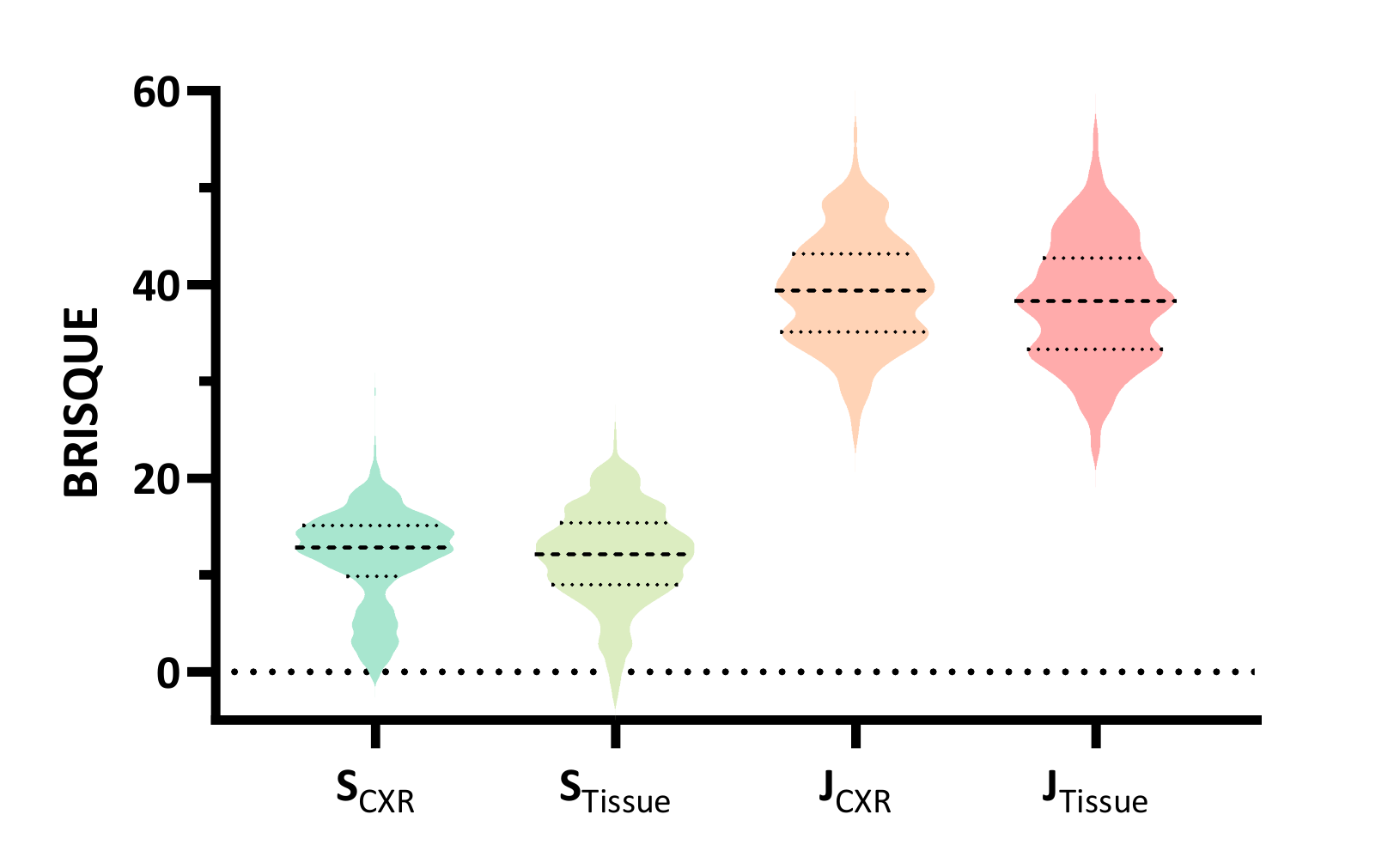}
    \caption{Violin plots of data quality using Blind/Referenceless Image Spatial Quality Evaluator (BRISQUE)~\cite{mittal2012no} ($\downarrow$) across two datasets (\textbf{S}: SZCH-X-Rays, \textbf{J}: JSRT~\cite{shiraishi2000development}). Our SZCH-X-Rays demonstrates higher quality with lower BRISQUE scores for both CXR and soft tissue data.}
    \label{IQA}
\end{figure}

In recent years, diffusion-based generative models~\cite{cao2026cross,song2025attridiffuser, ho2020denoising} have emerged as a powerful paradigm for medical image generation, effectively mitigating the mode collapse and convergence issues commonly associated with GANs. These models produce high-fidelity images through an iterative denoising process, rendering them particularly suitable for complex imaging tasks such as CXR bone suppression. Existing studies~\cite{sun2025gl, sun2025bs} have shown that diffusion-generated soft tissue images often exhibit superior perceptual quality compared with alternative generative approaches, thereby indicating improved visual plausibility and enhanced suitability for clinical interpretation. An early diffusion-based framework, BS-Diff~\cite{chen2024bs}, pioneered the application of Denoising Diffusion Probabilistic Models (DDPMs)~\cite{ho2020denoising} to bone suppression, leveraging a conditional diffusion architecture with a U-Net backbone and an AE-based enhancement module. Subsequently, BS-LDM~\cite{sun2025bs} extended this paradigm to Latent Diffusion Models (LDMs)~\cite{rombach2022high}, which operate in a compressed latent space to reduce computational overhead while enabling high-resolution image synthesis. Further advancing this line of research, GL-LCM~\cite{sun2025gl} introduced dual-path sampling and global-local fusion strategies, explicitly addressing the critical trade-off between bone suppression and detail retention. Despite these advancements, research in this field remains constrained by three core challenges:

\begin{itemize}

    \item While LDMs are widely adopted to handle high-resolution CXRs efficiently, their vision tokenizers are narrowly optimized for image reconstruction fidelity and therefore  fail to explicitly preserve structural consistency, which is important for distinguishing overlapping bony and soft tissue structures.

    \item Existing diffusion-based methods formulate bone suppression as a conditional generation task initialized from Gaussian noise, as illustrated in Fig.~\ref{paradigms} (a). The resulting distribution gap increases the difficulty of learning anatomically plausible mappings and often requires extensive iterative denoising.
    
    \item The scarcity of large-scale, high-quality paired datasets remains a primary constraint in bone suppression research. The largest existing publicly available dataset JSRT~\cite{shiraishi2000development} contains only 241 paired images with blurring and pronounced artifacts, as depicted in Figs.~\ref{IQA} and~\ref{oursvsgt}, limiting training reliability and clinical translation~\cite{schiller2025xu, rani2022multi}.
    
\end{itemize}

\begin{figure}[!t]
    \centering
    \includegraphics[width=\linewidth]{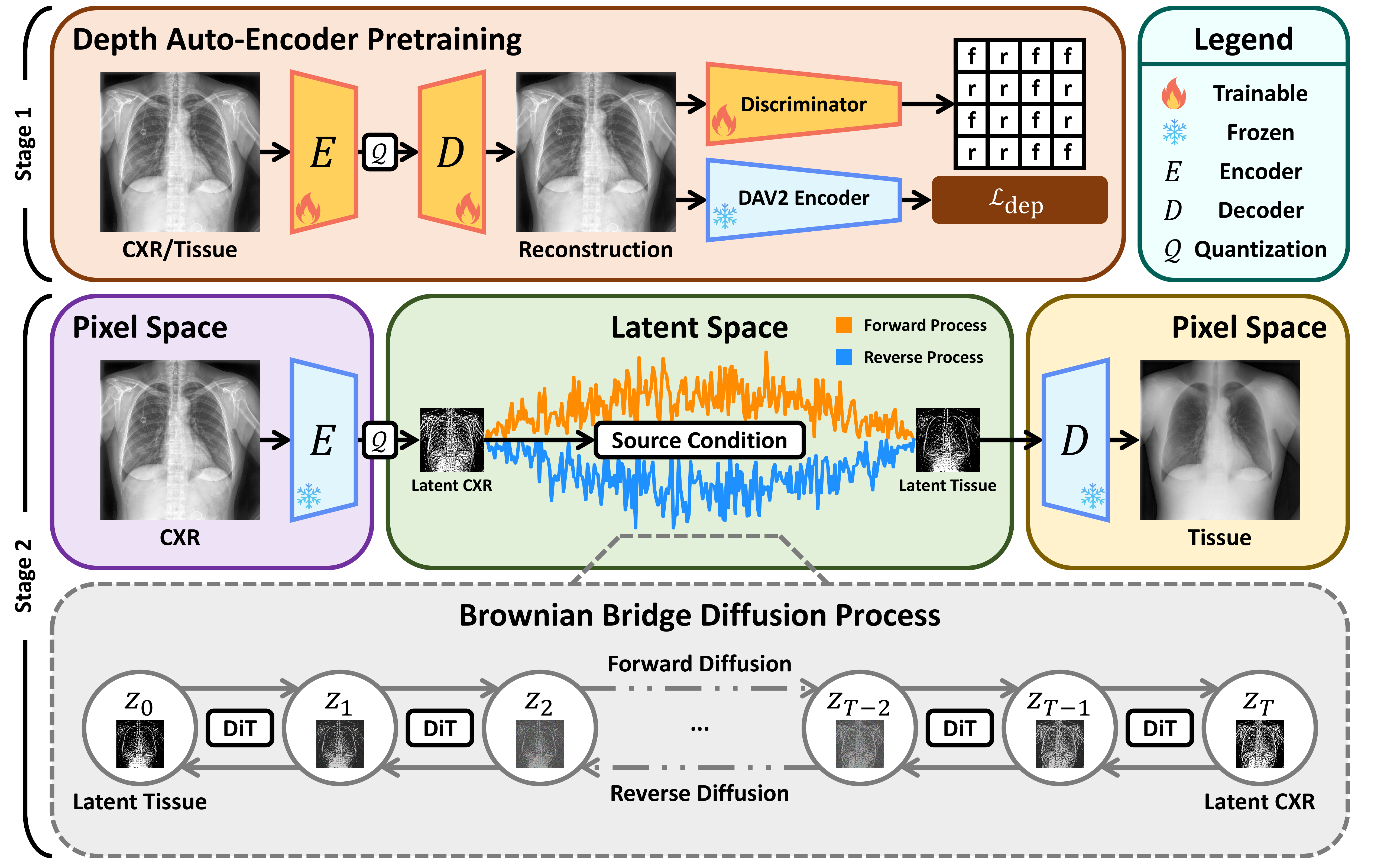}
    \caption{Overview of the DeBoneDiT architecture. Stage 1 (top): A Depth Auto-Encoder (DAE) is first pretrained to reconstruct images while preserving structural consistency, where a multi-level depth feature loss $\mathcal{L}_{\text{dep}}$ is computed between the input and reconstruction using a pretrained Depth Anything V2 (DAV2) encoder. Stage 2 (bottom): The bone suppression task is formulated as a source-conditioned Brownian bridge diffusion process in the latent space, where a DiT-based network iteratively transforms the latent representation of the source CXR $z_T$ into the target soft tissue $z_0$.}
    \label{frame}
\end{figure}

To address these challenges, we propose DeBoneDiT, a depth-driven conditional bridge diffusion Transformer architecture designed for efficient high-resolution bone suppression, as illustrated in Fig.~\ref{frame}. Drawing upon prior experience~\cite{sun2025gl, sun2025bs}, we adopt the two-stage design of LDMs to ensure computational efficiency. In addition, we have constructed and released SZCH-X-Rays, a high-quality dataset containing 741 pairs of CXR and DES soft tissue images for bone suppression research. Our contributions can be summarized as follows:

\begin{itemize}

    \item We introduce Depth Auto-Encoder (DAE), a structure-aware vision tokenizer for latent compression. By incorporating a multi-level depth feature loss derived from the pretrained Depth Anything V2~\cite{yang2024depth} encoder, DAE preserves both visual fidelity and structural consistency.

    \item To the best of our knowledge, we are the first to formulate bone suppression as a Brownian bridge diffusion process~\cite{li2023bbdm} aiming to narrow the source-to-target domain gap. Furthermore, we introduce a source conditioning mechanism that leverages CXRs as both the diffusion prior and structural guidance to maintain anatomical plausibility.

    \item We have constructed and released SZCH-X-Rays, the largest publicly available high-quality paired dataset for bone suppression to date, comprising 741 pairs of posterior-anterior CXR and DES soft tissue images.

    \item Extensive experiments and downstream evaluation demonstrate that DeBoneDiT achieves superior bone suppression performance with reduced computational overhead.

\end{itemize}

\section{Methodology}

As illustrated in Fig.~\ref{frame}, DeBoneDiT is a two-stage framework for high-resolution bone suppression. In the first stage, we introduce Depth Auto-Encoder (DAE) to compress high-dimensional images into latent representations, incorporating a multi-level depth feature loss to preserve structural consistency. In the second stage, we formulate the bone suppression task as a Brownian bridge diffusion process by establishing a direct bridge between the source CXR distribution and the target soft tissue distribution, effectively narrowing the inter-domain gap. This process is further enhanced by Source Condition Injection (SCI), which leverages the source CXR as explicit structural guidance to mitigate anatomical drift and error accumulation.

\subsection{Depth Auto-Encoder}

To explicitly preserve structural information during latent compression, we propose Depth Auto-Encoder (DAE) as a structure-aware vision tokenizer. Unlike conventional tokenizers that focus solely on surface-level reconstruction fidelity, DAE incorporates a multi-level depth feature loss derived from the pretrained Depth Anything V2 encoder, enabling it to preserve both visual fidelity and structural consistency.

Specifically, the DAE comprises an encoder $E$, a decoder $D$, a codebook $\mathcal{Z}$, a discriminator $\mathcal{D}_{\text{adv}}$, and a pretrained depth feature extractor $\phi$. Given an input CXR or soft tissue image $x \in \mathbb{R}^{1 \times H \times W}$, the encoder maps it into a continuous latent representation $z_e = E(x) \in \mathbb{R}^{C \times \frac{H}{r} \times \frac{W}{r}}$, where $r$ denotes the downsampling factor. Subsequently, $z_e$ is quantized against the codebook $\mathcal{Z}$ to obtain the discrete latent representation $z = \mathcal{Q}(z_e) \in \mathbb{R}^{C \times \frac{H}{r} \times \frac{W}{r}}$, where $\mathcal{Q}(\cdot)$ denotes the vector quantization operator. The decoder then reconstructs the image $\hat{x} = D(z)$. Furthermore, the discriminator $\mathcal{D}_{\text{adv}}$ distinguishes real from reconstructed images to drive the recovery of sharp, high-fidelity details.

Following the established latent space construction~\cite{sun2025bs, sun2025gl}, the objective function of DAE combines a composite reconstruction loss $\mathcal{L}_{\text{rec}}$ and supplementary constraints:
\begin{equation}
\mathcal{L}_{\text{DAE}} = \mathcal{L}_{\text{rec}} + \mathcal{L}_{\text{qua}} + \lambda_{\text{adv}} \mathcal{L}_{\text{adv}},
\end{equation}
where $\mathcal{L}_{\text{qua}}$ and $\mathcal{L}_{\text{adv}}$ represent the standard vector quantization loss~\cite{esser2021taming} and multi-scale adversarial loss~\cite{wang2018high}, respectively. Specifically, we define the reconstruction loss by integrating the depth feature loss with traditional pixel-wise and perceptual constraints:
\begin{equation}
\mathcal{L}_{\text{rec}} = \mathcal{L}_{\text{MSE}} + \lambda_{\text{per}} \mathcal{L}_{\text{per}} + \lambda_{\text{dep}} \mathcal{L}_{\text{dep}},
\end{equation}
where $\mathcal{L}_{\text{MSE}}$ and $\mathcal{L}_{\text{per}}$ denote the Mean Squared Error (MSE) loss and VGG-16-based perceptual loss~\cite{johnson2016perceptual}, respectively. 

\begin{figure}[!t]
    \centering
    \includegraphics[width=\linewidth]{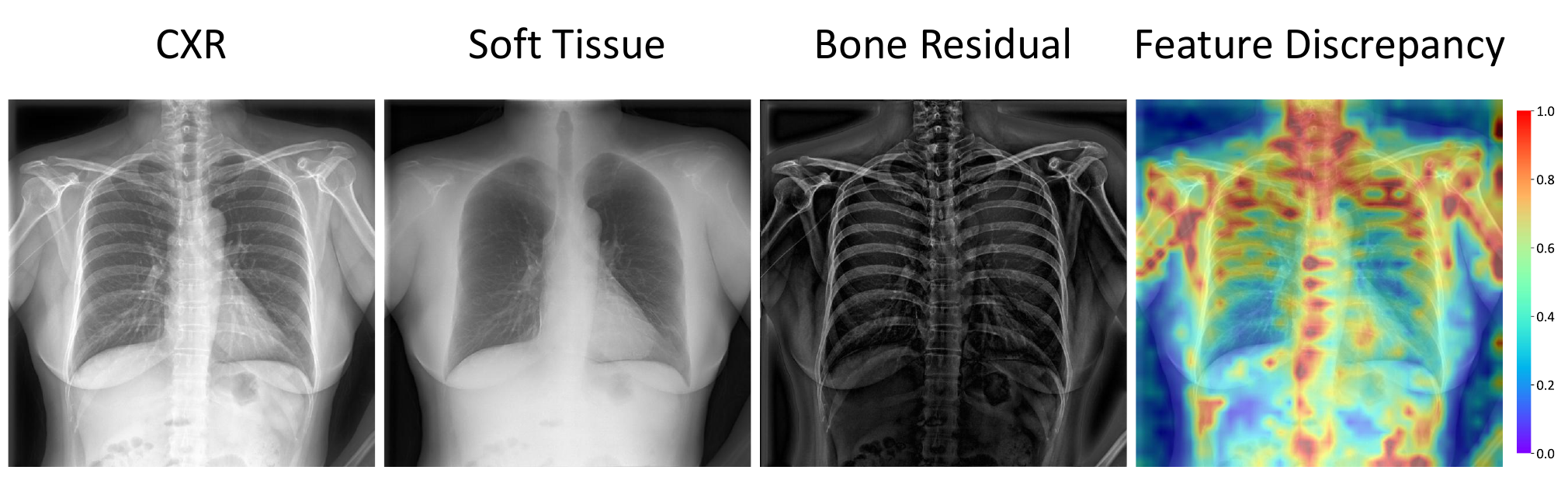}
    \caption{Visualization of bone residuals and depth feature discrepancies between a CXR and its corresponding soft tissue image. The depth feature difference map exhibits prominent responses around bony structures, indicating the sensitivity of depth features to skeletal anatomy.}
    \label{dep_vis}
\end{figure}

The introduced depth feature loss $\mathcal{L}_{\text{dep}}$ measures the feature discrepancy between the input and reconstructed images across multiple layers of the depth feature extractor $\phi$: 
\begin{equation}
\mathcal{L}_{\text{dep}} = \frac{1}{L} \sum_{l=1}^{L} \frac{1}{N \times D_l} \|\phi^l(x) - \phi^l(\hat{x})\|_F^2, 
\end{equation}
where $\phi^l(\cdot)$ extracts depth features from the $l$-th layer. Specifically, for each selected layer $l$, $\phi^l(x)$ has a shape of $N_\text{p} \times D_l$, where $N_\text{p}$ denotes the sequence length of spatial patch tokens, and $D_l$ denotes the channel dimension of the $l$-th layer. In this work, we utilize the pretrained Depth Anything V2~\cite{yang2024depth} encoder as the depth feature extractor $\phi$. Rather than predicting explicit depth maps, we leverage its structure-aware features to constrain latent reconstruction, as visualized in Fig.~\ref{dep_vis}.

Following the conventions established in previous studies~\cite{sun2025bs, sun2025gl}, the trade-off weights for $\mathcal{L}_{\text{adv}}$ and $\mathcal{L}_{\text{per}}$ are detailed in Section~\ref{sec:id}. Regarding the newly introduced depth loss $\mathcal{L}_{\text{dep}}$, a comprehensive sensitivity analysis for $\lambda_{\text{dep}}$ is provided in Section~\ref{sec:iotdl}. 

\subsection{Brownian Bridge Diffusion Process}

Rather than initiating the generation process from pure Gaussian noise, we formulate bone suppression as a Brownian bridge diffusion process~\cite{li2023bbdm}, leveraging CXRs as the diffusion prior. By modeling the transition directly between the source CXR and target soft tissue domains, this approach effectively preserves the structural correspondence between the CXR and synthesized soft tissue.

Let $z_T=y$ and $z_0$ denote the latent representations of the source CXR and target soft tissue, respectively. We define a forward diffusion process that transitions $z_0$ to $z_T$ over $T$ timesteps. The intermediate state $z_t$ at timestep $t \in \{0, 1, \dots, T\}$ is derived from the following conditional distribution:
\begin{equation}
q(z_t | z_0, y) = \mathcal{N}(z_t; (1 - m_t)z_0 + m_t y, \delta_t \mathbf{I}),
\end{equation}
where $m_t = \frac{t}{T}$ and $\delta_t = 2(m_t-m_t^2)$ define the interpolation and variance schedules of the Brownian bridge, respectively. This formulation ensures that the mean of $z_t$ linearly interpolates between the two endpoints, while the stochastic variance vanishes at $t=0$ and $t=T$.

During the training phase, the model learns to reconstruct the target $z_0$ from the intermediate state $z_t$. Specifically, $z_t$ is sampled by perturbing the target $z_0$ with a mixture of the source $y$ and Gaussian noise $\epsilon \sim \mathcal{N}(\mathbf{0}, \mathbf{I})$:
\begin{equation}
z_t = (1 - m_t)z_0 + m_t y + \sqrt{\delta_t}\epsilon.
\end{equation}
In this work, we employ a denoising network $\epsilon_\theta$ based on the Diffusion Transformer (DiT)~\cite{peebles2023scalable} to predict the added noise $\epsilon$. The training objective of the Brownian bridge diffusion process $\mathcal{L}_{\text{BBDP}}$ is defined as:
\begin{equation}
\label{eq_training_obj}
\mathcal{L}_{\text{BBDP}} = \mathbb{E}_{t, z_0, y, \epsilon}\lVert\epsilon - \epsilon_\theta(z_t, t)\rVert^2_2,
\end{equation}
where $t \sim \mathcal{U}\{0, 1, \dots, T-1\}$, and $(z_0, y) \sim q(z_0, y)$. 

In the reverse diffusion process, the model iteratively reconstructs the target soft tissue $z_0$ initialized from the source CXR $ z_T = y$. 
The target representation $\hat{z}_0$ at any timestep $t$ can be dynamically estimated from the current state $z_t$ as:
\begin{equation}
\label{eq_z0}
\hat{z}_0 = \frac{1}{1-m_t} (z_t - m_t y - \sqrt{\delta_t} \epsilon_\theta(z_t, t)).
\end{equation}
Based on the estimated target representation $\hat{z}0$, we define the reverse transition distribution for sampling the previous state $z{t-1}$ as:
\begin{equation}
p_\theta(z_{t-1} | z_t, \hat{z}_0, y) = \mathcal{N}(z_{t-1}; (1 - m_{t-1}) \hat{z}_0 + m_{t-1} y, \tilde{\delta}_t \mathbf{I}).
\end{equation}
The state at the previous timestep $t-1$ is sampled as:
\begin{equation}
z_{t-1} = (1 - m_{t-1}) \hat{z}_0 + m_{t-1} y + \sqrt{\tilde{\delta}_t} \xi,
\end{equation}
where $\xi \sim \mathcal{N}(\mathbf{0}, \mathbf{I})$ if $t > 1$ (else $\xi = \mathbf{0}$), and the reverse step variance $\tilde{\delta}_t$ is derived from the forward schedules via Bayes' rule \cite{li2023bbdm}. This iterative refinement guides the progressive translation from the CXR domain to the soft tissue domain, enabling the gradual removal of overlapping bony structures.

\subsection{Source Condition Injection}

While the vanilla Brownian bridge diffusion process incorporates the source CXR $y$ as a boundary constraint, relying solely on this implicit formulation leads to anatomical drift and error accumulation during sampling. To provide explicit, step-wise structural reinforcement, we introduce an explicit source conditioning mechanism, SCI. Specifically, at each denoising step, we concatenate the noisy latent $z_t$ with the latent representation of the source CXR $y$ along the channel dimension before feeding them into the denoising network $\epsilon_\theta$. Accordingly, the training objective in Eq.~\ref{eq_training_obj} is reformulated as:
\begin{equation}
\mathcal{L}_{\text{BBDP}} = \mathbb{E}_{t, z_0, y, \epsilon}\lVert\epsilon - \epsilon_\theta(z_t, t, y)\rVert^2_2.
\end{equation}
This explicit conditioning enables the denoising network $\epsilon_\theta$ to leverage anatomical cues from the source CXR at each step, ensuring more faithful soft tissue synthesis.

Accordingly, this modification extends to the reverse diffusion process, where the one-step estimation of the target representation $\hat{z}_0$ in Eq.~\ref{eq_z0} is updated to incorporate the explicit source guidance:
\begin{equation}
\hat{z}_0 = \frac{1}{1-m_t} (z_t - m_t y - \sqrt{\delta_t} \epsilon_\theta(z_t, t, y)).
\end{equation}
By leveraging the source CXR $y$ as both a boundary constraint in the sampling equation and a conditional input to the network $\epsilon_\theta$ during the iterative sampling process, the model further mitigates anatomical drift and error accumulation. The overall training and sampling processes are summarized in Algorithm~\ref{al_1} and Algorithm~\ref{al_2}, respectively.

\begin{algorithm}[!h]
\caption{Training}
\label{al_1}
\begin{algorithmic}[1]
\REPEAT
    \STATE Paired latent data $(z_0, y) \sim q(z_0, y)$
    \STATE Timestep $t \sim \mathcal{U}\{0, 1, \dots, T-1\}$
    \STATE Gaussian noise $\epsilon \sim \mathcal{N}(\mathbf{0}, \mathbf{I})$
    \STATE Forward diffusion $z_t = (1 - m_t)z_0 + m_t y + \sqrt{\delta_t}\epsilon$
    \STATE Take gradient descent step on
    \STATE \hskip 2em $\nabla_\theta \lVert\epsilon - \epsilon_\theta(z_t, t, y)\rVert^2_2$
\UNTIL{converged}
\end{algorithmic}
\end{algorithm}

\begin{algorithm}[!h]
\caption{Sampling}
\label{al_2}
\begin{algorithmic}[1]
\STATE Sample conditional input $z_T = y \sim q(y)$
\STATE Set $z_{T-1} = z_T$
\FOR{$t = T-1, \dots, 1$}
    \STATE $\xi \sim \mathcal{N}(\mathbf{0}, \mathbf{I})$ if $t > 1$, else $\xi = \mathbf{0}$
    \STATE $\hat{z}_0 = \frac{1}{1-m_t}(z_t - m_t y - \sqrt{\delta_t} \epsilon_\theta(z_t, t, y))$
    \STATE $z_{t-1} = (1-m_{t-1}) \hat{z}_0 + m_{t-1} y + \sqrt{\tilde{\delta}_t} \xi$
\ENDFOR
\STATE \textbf{return} $z_0$
\end{algorithmic}
\end{algorithm}

\section{Experiments and Discussion}

\subsection{Datasets and Evaluation metrics}

\subsubsection{Data Preparation}

We conducted comprehensive experiments across three distinct datasets, with each dataset assigned a specific role in evaluation: our self-constructed SZCH-X-Rays dataset and the publicly available JSRT dataset~\cite{shiraishi2000development} were utilized for performance evaluation, while the Asraf dataset~\cite{aljuaid2026experimental} was employed for downstream evaluation to benchmark clinical applicability. All the images were resized to 1024 $\times$ 1024 pixels for experimental consistency. 

\textbf{SZCH-X-Rays} comprises 741 pairs of posterior-anterior CXR and DES soft tissue images, acquired using a GE Discovery XR656 system in collaboration with our partner hospital. Initially stored in 14-bit DICOM format, the images were converted to PNG format to streamline the processing workflow, which feature an original resolution of $2021 \times 2021$ with a pixel size of 0.194 mm. Consistent with prior studies~\cite{bae2022bone, yang2017cascade}, examinations affected by operational errors, pronounced motion artifacts, severe pleural effusion, or pneumothorax were excluded because these conditions may substantially degrade the quality of DES soft tissue references. Finally, the dataset was partitioned into 592 training, 74 validation and 75 test pairs. The dataset is publicly available under the CC BY-NC 4.0 license.

\textbf{JSRT} contains 241 pairs of CXR and synthetic soft tissue images, split into 192 training, 24 validation and 25 test pairs. The CXR images were sourced from the Japanese Society of Radiological Technology while the corresponding soft tissue images were algorithmically synthesized by researchers at the Budapest University of Technology and Economics. The original images feature a resolution of $2048 \times 2048$ pixels.

\textbf{Asraf} encompasses 6939 posterior-anterior CXR images sourced from publicly available resources, evenly distributed across three diagnostic classes with 2313 images each: pneumonia, COVID-19, and normal. Original resolutions vary from $205 \times 138$ to $5623 \times 4757$ pixels, with $1024 \times 1024$ being the most prevalent size at 45.2\%. To prevent distortion in the non-square images, we cropped each to its maximum square. To evaluate clinical applicability, we performed downstream classification on this dataset using a 5-fold cross-validation strategy to ensure robustness.

\subsubsection{Evaluation Metrics}
We employ five main quantitative metrics to evaluate the bone suppression performance, including Learned Perceptual Image Patch Similarity (LPIPS)~\cite{zhang2018unreasonable} ($\downarrow$), Mean Squared Error (MSE) ($\downarrow$), Peak Signal-to-Noise Ratio (PSNR) ($\uparrow$), Structural Similarity Index Measure (SSIM) ($\uparrow$), and Bone Suppression Ratio (BSR)~\cite{hogeweg2013suppression} ($\uparrow$), a metric that specifically measures the proportion of bone signal effectively suppressed relative to the original bone energy. For downstream task evaluation, we report four classification metrics, namely accuracy ($\uparrow$), precision ($\uparrow$), recall ($\uparrow$), and F1-score ($\uparrow$). Additionally, we utilize the Blind/Referenceless Image Spatial Quality Evaluator (BRISQUE)~\cite{mittal2012no} ($\downarrow$) to evaluate the quality of both generated images and ground truth.

\subsection{Implementation Details}
\label{sec:id}
All experiments were performed using PyTorch 2.5.1 on an NVIDIA A800 GPU within Ubuntu 22.04. The DAE was trained from scratch for 200 epochs with a batch size of 4 using the Adam optimizer. Specifically, the learning rate was set to $10^{-4}$ for both the encoder $E$ and decoder $D$, and $5 \times 10^{-4}$ for the discriminator $\mathcal{D}_{\text{adv}}$. Regarding the latent space, the downsampling factor $r$ and the channel dimension $C$ were set to 4, resulting in a latent representation of $z \in \mathbb{R}^{4 \times 256 \times 256}$ for input $x \in \mathbb{R}^{1 \times 1024 \times 1024}$. The codebook $\mathcal{Z}$ was configured with a size of $K = 1024$. For the depth feature extractor $\phi$, we adopted the ViT-B/14 configuration, with input data resized to $518 \times 518$ pixels prior to depth feature extraction. Consequently, the sequence length of spatial patch tokens was $N_\text{p} = 1369$, and the feature dimensions $D_l$ for the $L = 4$ selected intermediate layers were 96, 192, 384, and 768, respectively. To balance different training objectives, the trade-off weights for DAE training were set to $\lambda_{\text{adv}}=10^{-2}$, $\lambda_{\text{per}}=10^{-3}$, and $\lambda_{\text{dep}}=1$. 

In addition, the DiT-based denoising network $\epsilon_\theta$ with $N = 12$ DiT blocks was trained from scratch for 2000 epochs with a batch size of 8 using the AdamW optimizer. Specifically, a multi-step learning rate decay strategy was employed, initialized at $10^{-4}$. Following established practices~\cite{sun2025bs}, Exponential Moving Average (EMA) and Classifier-Free Guidance (CFG)~\cite{ho2022classifier} were adopted to ensure model robustness during training and inference, respectively. Furthermore, the number of diffusion timesteps $T$ was set to 1000 for training, whereas only 50 sampling steps were utilized for inference to facilitate rapid generation. 

For the downstream evaluation, all five classification models were fine-tuned from ImageNet-1K~\cite{russakovsky2015imagenet} pretrained weights using the Adam optimizer with a weight decay of $10^{-4}$. The initial learning rate was set to $10^{-3}$ with a plateau-based learning rate scheduler. All these models were trained for 20 epochs with a batch size of 16 using the standard cross-entropy loss with a label smoothing coefficient of 0.15.

\subsection{Main Results}

We conduct a rigorous quantitative evaluation against seven state-of-the-art methods, encompassing conventional GAN-based Yang et al.~\cite{yang2025improving}, MDA-GAN~\cite{singh2024mda} and Pix2Pix~\cite{rani2022multi}, the U-Net-based xU-NetFullSharp~\cite{schiller2025xu}, and recent diffusion-based GL-LCM~\cite{sun2025gl}, BS-LDM~\cite{sun2025bs} and BS-Diff~\cite{chen2024bs}. Table~\ref{table_comp1} presents the performance on both our SZCH-X-Rays and the public JSRT datasets. DeBoneDiT consistently outperforms all competing methods across all five metrics, demonstrating its superior capability in bone suppression and detail preservation.

\begin{table}[!t]
\setlength{\tabcolsep}{2pt}
\caption{Quantitative comparison of different methods on the SZCH-X-Rays and JSRT datasets. Statistical significance compared to the second-best counterpart is evaluated using the Wilcoxon signed-rank test with Holm-Bonferroni correction applied independently for each dataset ($^{*}: p < 0.05$, $^{**}: p < 0.01$, $^{***}: p < 0.001$).}\label{table_comp1}
\centering
\resizebox{\linewidth}{!}{
\begin{tabular}{p{2.4cm}<{\centering}p{2.7cm}<{\centering}p{1.8cm}<{\centering}p{2.5cm}<{\centering}p{2.5cm}<{\centering}p{2.5cm}<{\centering}p{2.5cm}<{\centering}p{2.5cm}<{\centering}}
\toprule
Dataset & Method & Source & LPIPS ($\downarrow$) & MSE ($10^{-3}$, $\downarrow$) & PSNR (dB, $\uparrow$) & SSIM (\%, $\uparrow$) &  BSR (\%, $\uparrow$) \\
\midrule
\multirow{8}{*}{SZCH-X-Rays}
& xU-NetFullSharp & BSPC$_{25}$ & 0.141 $\pm$ 0.032 & 2.154 $\pm$ 1.544 & 27.885 $\pm$ 3.418 & 87.269 $\pm$ 3.685 & 91.934 $\pm$ 5.482\\
& Pix2Pix & ISWA$_{22}$ & 0.196 $\pm$ 0.039 & 1.503 $\pm$ 0.812 & 28.886 $\pm$ 2.460 & 83.884 $\pm$ 3.022 & 92.842 $\pm$ 3.816\\
& MDA-GAN & TAI$_{24}$ & 0.141 $\pm$ 0.033 & 1.866 $\pm$ 1.389 & 28.549 $\pm$ 3.464 & 87.854 $\pm$ 4.080 & 92.634 $\pm$ 4.908\\
& Yang et al. & TMI$_{25}$ & 0.203 $\pm$ 0.039 & 1.392 $\pm$ 0.592 & 28.953 $\pm$ 1.855 & 78.105 $\pm$ 2.797 & 92.349 $\pm$ 3.153\\
& BS-Diff & ISBI$_{24}$ & 0.189 $\pm$ 0.038 & 1.975 $\pm$ 1.137 & 27.897 $\pm$ 2.921 & 87.673 $\pm$ 4.005 & 92.821 $\pm$ 4.980\\
& BS-LDM & JBHI$_{25}$ & 0.103 $\pm$ 0.016 & 1.615 $\pm$ 0.933 & 28.587 $\pm$ 2.435 & 78.412 $\pm$ 2.861 & 92.173 $\pm$ 4.314\\
& GL-LCM & MICCAI$_{25}$ & 0.093 $\pm$ 0.019 & 1.977 $\pm$ 1.403 & 28.126 $\pm$ 3.210 & 80.716 $\pm$ 2.923 & 92.814 $\pm$ 4.198\\
& \textbf{DeBoneDiT} & \textbf{Ours} & \textbf{0.078 $\pm$ 0.023}$^{***}$ & \textbf{1.233 $\pm$ 0.626}$^{**}$ & \textbf{29.620 $\pm$ 2.168}$^{***}$ & \textbf{88.204 $\pm$ 2.329}$^{*}$ & \textbf{93.143 $\pm$ 4.160}$^{*}$\\
\midrule
\multirow{8}{*}{JSRT}
& xU-NetFullSharp & BSPC$_{25}$ & 0.096 $\pm$ 0.031 & 1.257 $\pm$ 0.758 & 29.883 $\pm$ 2.874 & 93.764 $\pm$ 2.053 & 91.525 $\pm$ 8.223\\
& Pix2Pix & ISWA$_{22}$ & 0.112 $\pm$ 0.034 & 1.119 $\pm$ 0.739 & 30.524 $\pm$ 3.123 & 93.162 $\pm$ 2.066 & 90.655 $\pm$ 9.135\\
& MDA-GAN & TAI$_{24}$ & 0.098 $\pm$ 0.031 & 1.368 $\pm$ 0.876 & 29.655 $\pm$ 3.069 & 93.839 $\pm$ 1.960 & 90.851 $\pm$ 8.065\\
& Yang et al. & TMI$_{25}$ & 0.083 $\pm$ 0.030 & 1.115 $\pm$ 0.787 & 30.706 $\pm$ 3.390 & 93.680 $\pm$ 1.895 & 90.815 $\pm$ 9.355\\
& BS-Diff & ISBI$_{24}$ & 0.128 $\pm$ 0.040 & 1.640 $\pm$ 1.050 & 28.841 $\pm$ 3.063 & 93.012 $\pm$ 2.470 & 91.023 $\pm$ 8.561\\
& BS-LDM & JBHI$_{25}$ & 0.063 $\pm$ 0.026 & 1.262 $\pm$ 0.867 & 29.881 $\pm$ 2.786 & 93.447 $\pm$ 1.759 & 88.733 $\pm$ 11.648\\
& GL-LCM & MICCAI$_{25}$ & 0.058 $\pm$ 0.024 & 1.222 $\pm$ 0.892 & 30.217 $\pm$ 3.119 & 93.685 $\pm$ 1.818 & 91.963 $\pm$ 9.231\\
& \textbf{DeBoneDiT} & \textbf{Ours} & \textbf{0.032 $\pm$ 0.013}$^{***}$ & \textbf{1.017 $\pm$ 0.983}$^{*}$ & \textbf{31.339 $\pm$ 3.280}$^{**}$ & \textbf{94.262 $\pm$ 1.233}$^{*}$ & \textbf{93.530 $\pm$ 9.296}\\
\bottomrule
\end{tabular}
}
\end{table}

On the SZCH-X-Rays dataset, DeBoneDiT establishes a new state-of-the-art, surpassing the second-best baseline by 0.667 dB in PSNR and achieving a superior SSIM of 88.204\%. Notably, our method yields the lowest LPIPS of 0.078, demonstrating enhanced perceptual fidelity compared to other methods. This advantage is even more pronounced on the JSRT dataset, where DeBoneDiT achieves a remarkably low LPIPS of 0.032, which represents a significant improvement over the 0.058 of the second-best GL-LCM, alongside a robust BSR of 93.530\%. These results validate the effectiveness and consistency of our approach across diverse data distributions.

\begin{figure}[!t]
    \centering
    \includegraphics[width=\linewidth]{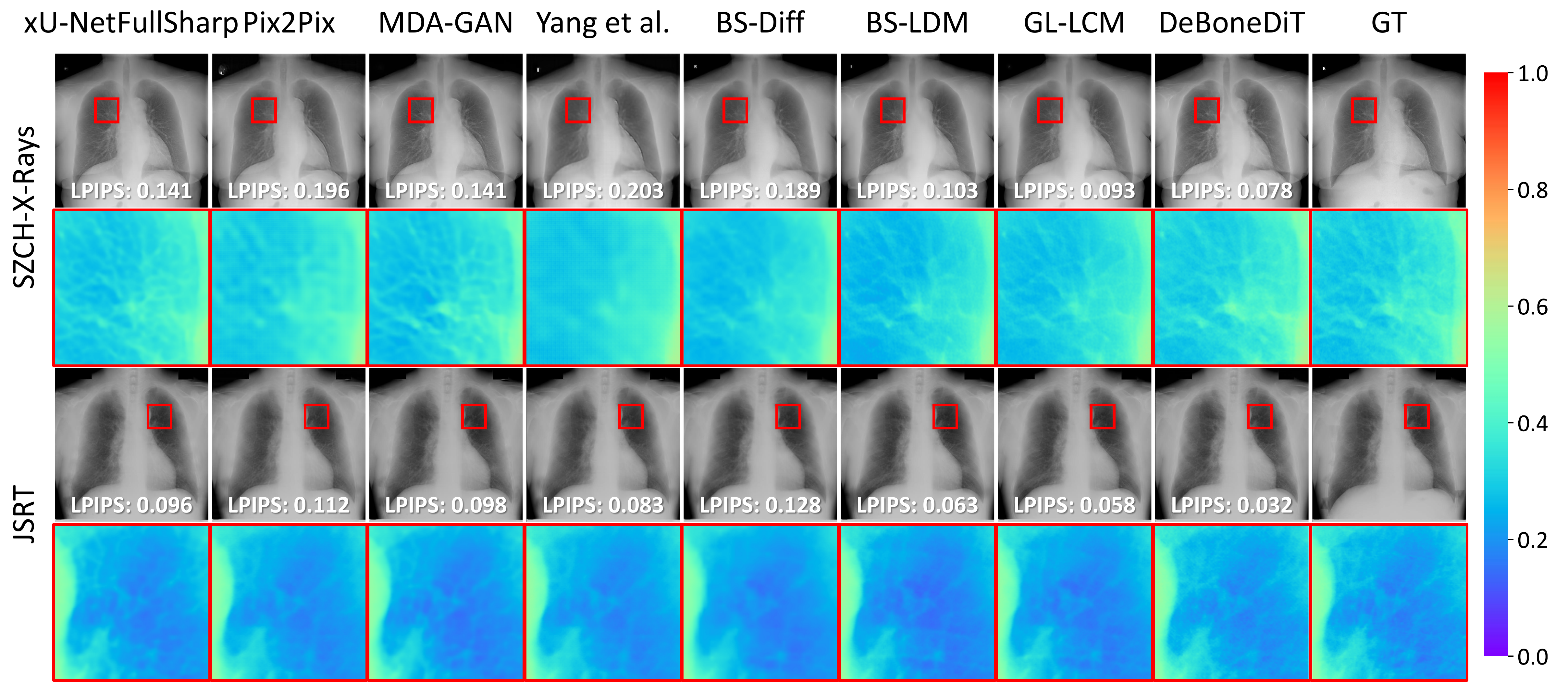}
    \caption{Visual comparison of different methods on the SZCH-X-Rays and JSRT datasets. The even rows display pseudo-color magnified views of the pixel intensities within the regions indicated by red boxes to highlight subtle textural details. (GT: Ground Truth)}
    \label{comp_fig}
\end{figure}

\begin{figure}[!t]
    \centering
    \includegraphics[width=\linewidth]{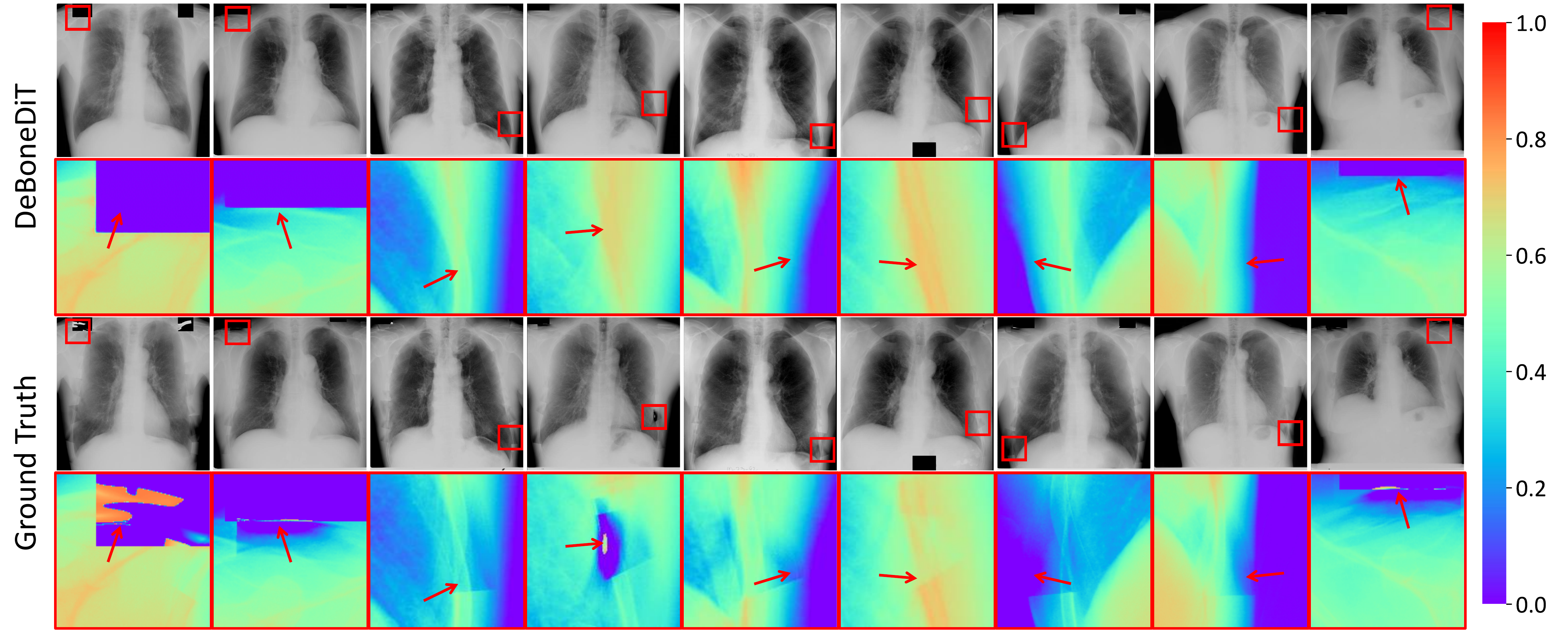}
    \caption{Qualitative comparison of the outputs of DeBoneDiT (top) against the ground truth (bottom) from JSRT. As indicated by the red arrows, the ground truth contains visible acquisition artifacts. In contrast, DeBoneDiT produces anatomically coherent soft tissue structures without reproducing these artifacts. This observation indicates a degree of robustness to imperfections in the reference images and suggests that the model does not simply memorize artifact patterns from the training data.}
    \label{oursvsgt}
\end{figure}

Beyond pixel-level metrics, we evaluate the visual fidelity of the synthesized soft tissue images, as illustrated in Fig.~\ref{comp_fig}. Qualitative comparisons reveal that GAN-based frameworks, exemplified by Pix2Pix~\cite{rani2022multi}, and early diffusion-based approaches, such as BS-Diff~\cite{chen2024bs}, tend to over-smooth fine pulmonary structures. While more recent methods like BS-LDM~\cite{sun2025bs} improve detail preservation, they exhibit performance instability across different data distributions. Specifically, while demonstrating adequate suppression on SZCH-X-Rays, they frequently encounter residual bone shadows on JSRT. In contrast, DeBoneDiT consistently and effectively suppresses overlapping bony structures while preserving fine pulmonary details across both datasets. Furthermore, Fig.~\ref{oursvsgt} suggests that DeBoneDiT does not simply memorize artifact patterns from the training data, indicating a degree of robustness to imperfections in the reference images.

\begin{table}[!t]
\caption{Quantitative comparison of model complexity and inference latency among different diffusion-based methods.}\label{table_inf_eff}
\centering
\begin{tabular}{p{1.9cm}<{\centering}p{1.9cm}<{\centering}p{1.9cm}<{\centering}p{1.9cm}<{\centering}p{1.9cm}<{\centering}}
\toprule
Method & Scheduler & Timesteps & Params (M) & Time (s)\\
\midrule
BS-Diff & DDPM & 1000 & 254.7 & 110.5\\
BS-LDM & DDPM & 1000  & 421.3 & 85.9\\
GL-LCM & LCM & \textbf{50}  & 436.9 & 8.7\\
\textbf{DeBoneDiT} & \textbf{BBDM} & \textbf{50}  & \textbf{72.8} & \textbf{3.4}\\
\bottomrule
\end{tabular}
\end{table}

Regarding efficiency, Table~\ref{table_inf_eff} compares the model complexity and inference latency of various diffusion-based frameworks. By leveraging the Brownian bridge diffusion process, DeBoneDiT achieves a rapid inference time of 3.4 seconds per image. Moreover, with a compact parameter count of only 72.8 M, our DiT-based architecture is more lightweight than traditional U-Net-based counterparts. This superior efficiency underscores the potential of DeBoneDiT as a scalable and viable solution for high-throughput clinical workflows.

\subsection{Ablation Study}

\begin{table}[!t]
\caption{Quantitative ablation study on the impact of the depth feature loss $\mathcal{L}_{\text{dep}}$.}\label{table_abl_dep}
\centering
\begin{tabular}{p{2.4cm}<{\centering}p{1.2cm}<{\centering}p{2.4cm}<{\centering}p{2.4cm}<{\centering}}
\toprule
Dataset & $\mathcal{L}_{\text{dep}}$ & LPIPS ($\downarrow$) & BRISQUE ($\downarrow$)\\
\midrule
\multirow{2}{*}{SZCH-X-Rays}
& $\times$ & 0.169 $\pm$ 0.067 & 26.644 $\pm$ 2.591 \\
& $\checkmark$ & \textbf{0.078 $\pm$ 0.023} & \textbf{22.043 $\pm$ 5.001} \\
\midrule
\multirow{2}{*}{JSRT}
& $\times$ & 0.061 $\pm$ 0.023 & 47.003 $\pm$ 0.873 \\
& $\checkmark$ & \textbf{0.032 $\pm$ 0.013} & \textbf{37.144 $\pm$ 3.787} \\
\bottomrule
\end{tabular}
\end{table}

\begin{figure}[!t]
    \centering
    \includegraphics[width=\linewidth]{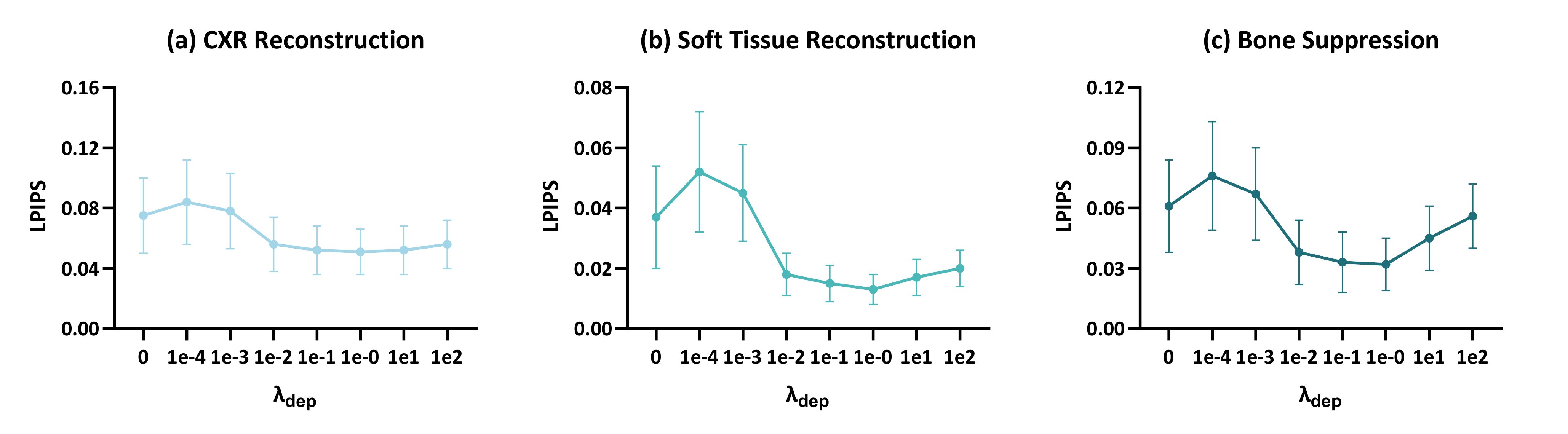}
    \caption{Sensitivity analysis of the depth feature loss weight $\lambda_{\text{dep}}$ on JSRT. The results indicate that setting $\lambda_{\text{dep}}=1$ attains the minimal perceptual error across (a) CXR reconstruction, (b) soft tissue reconstruction, and (c) bone suppression, demonstrating its optimal balance for the task.}
    \label{dep}
\end{figure}

\begin{figure}[!t]
    \centering
    \includegraphics[width=0.8\linewidth]{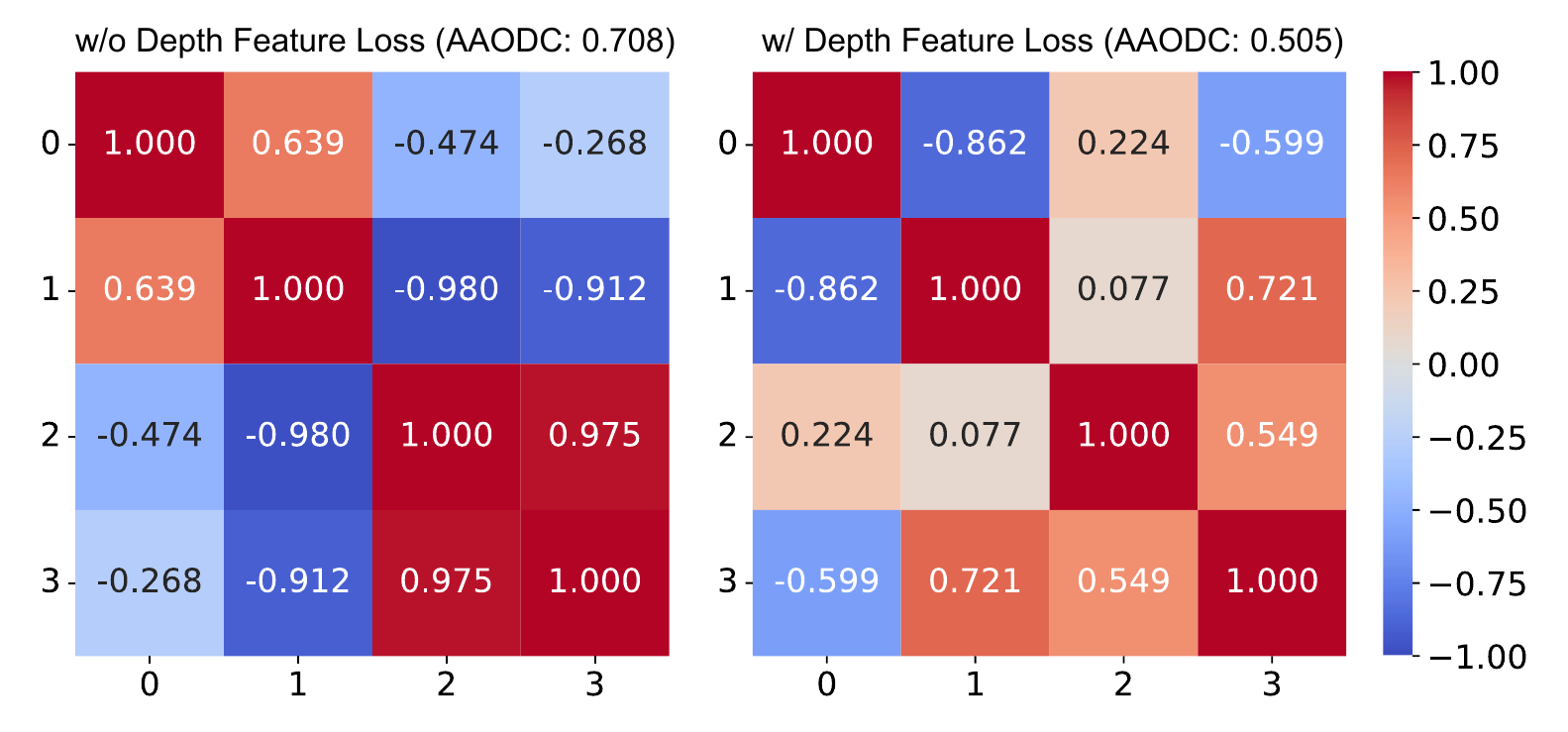}
    \caption{Latent channel correlation analysis of DAE on SZCH-X-Rays. A lower Average Absolute Off-Diagonal Correlation (AAODC)~\cite{zbontar2021barlow} indicates a higher degree of feature disentanglement. }
    \label{co}
\end{figure}

\subsubsection{Impact of the Depth Feature Loss}
\label{sec:iotdl}
We first evaluate the impact of the depth feature loss $\mathcal{L}_{\text{dep}}$ on capturing structural hierarchies. As quantified in Table~\ref{table_abl_dep}, incorporating $\mathcal{L}_{\text{dep}}$ significantly reduces LPIPS from 0.169 / 0.061 to 0.078 / 0.032, underscoring that depth-driven supervision can enhance the perceptual quality of bone-suppressed images. Specifically, the sensitivity analysis in Fig.~\ref{dep} demonstrates that $\lambda_{dep}=1$ provides the optimal regularization strength, minimizing perceptual distortion across both image reconstruction and bone suppression tasks. This improvement is further supported by the latent channel correlation analysis in Fig.~\ref{co}, where the inclusion of $\mathcal{L}_{\text{dep}}$ leads to a lower Average Absolute Off-Diagonal Correlation (AAODC)~\cite{zbontar2021barlow}, indicating superior feature disentanglement within the latent space.

\begin{table}[!t]
\setlength{\tabcolsep}{1pt}
\caption{Quantitative ablation study on the impact of different diffusion paradigms, including CDM, BBDM and DeBoneDiT (Ours).}\label{table_abl}
\centering
\resizebox{\linewidth}{!}{
\begin{tabular}{p{2.4cm}<{\centering}p{2cm}<{\centering}p{1.4cm}<{\centering}p{1.6cm}<{\centering}p{2.6cm}<{\centering}p{2.6cm}<{\centering}p{2.6cm}<{\centering}p{2.6cm}<{\centering}}
\toprule
Dataset & Paradigm & Prior & Condition & LPIPS ($\downarrow$) & MSE ($10^{-3}$, $\downarrow$) & PSNR (dB, $\uparrow$) & BSR (\%, $\uparrow$) \\
\midrule
\multirow{3}{*}{SZCH-X-Rays}
& CDM & $\mathcal{N}(\mathbf{0}, \mathbf{I})$ & $y$ & 0.093 $\pm$ 0.018 & 2.004 $\pm$ 1.207 & 27.835 $\pm$ 2.861 & 92.141 $\pm$ 4.031\\
& BBDM & $y$ & $\times$ & 0.137 $\pm$ 0.040 & 1.558 $\pm$ 0.829 & 28.602 $\pm$ 2.103 & 86.973 $\pm$ 10.154\\
& \textbf{DeBoneDiT} & $y$ & $y$ & \textbf{0.078 $\pm$ 0.023} & \textbf{1.233 $\pm$ 0.626} & \textbf{29.620 $\pm$ 2.168} & \textbf{93.143 $\pm$ 4.160}\\
\midrule
\multirow{3}{*}{JSRT}
& CDM & $\mathcal{N}(\mathbf{0}, \mathbf{I})$ & $y$ & 0.058 $\pm$ 0.024 & 1.212 $\pm$ 0.881 & 30.257 $\pm$ 3.138 & 91.792 $\pm$ 9.406\\
& BBDM & $y$ & $\times$ & 0.046 $\pm$ 0.015 & 2.021 $\pm$ 0.972 & 27.311 $\pm$ 1.665 & 90.409 $\pm$ 9.532\\
& \textbf{DeBoneDiT} & $y$ & $y$ & \textbf{0.032 $\pm$ 0.013} & \textbf{1.017 $\pm$ 0.983} & \textbf{31.339 $\pm$ 3.280} & \textbf{93.530 $\pm$ 9.296}\\
\bottomrule
\end{tabular}
}
\end{table}

\begin{figure}[!t]
    \centering
    \includegraphics[width=0.8\linewidth]{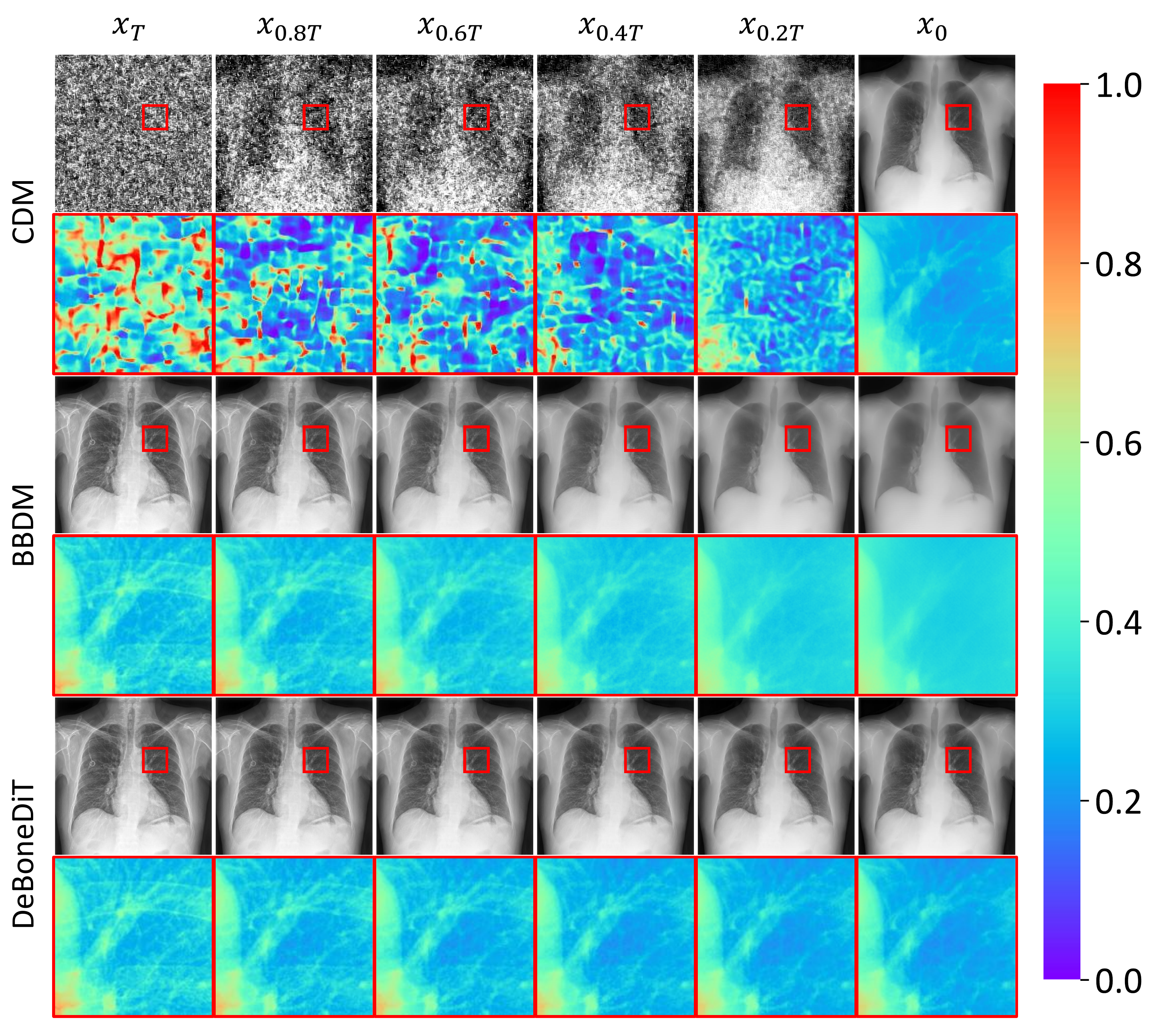}
    \caption{Sampling process comparison of three diffusion paradigms on SZCH-X-Rays. CDM is initialized from Gaussian noise, while BBDM and DeBoneDiT leverage the source CXR as the prior. Notably, DeBoneDiT demonstrates superior detail preservation compared to BBDM.}
    \label{process}
\end{figure}

\begin{figure}[!t]
    \centering
    \includegraphics[width=\linewidth]{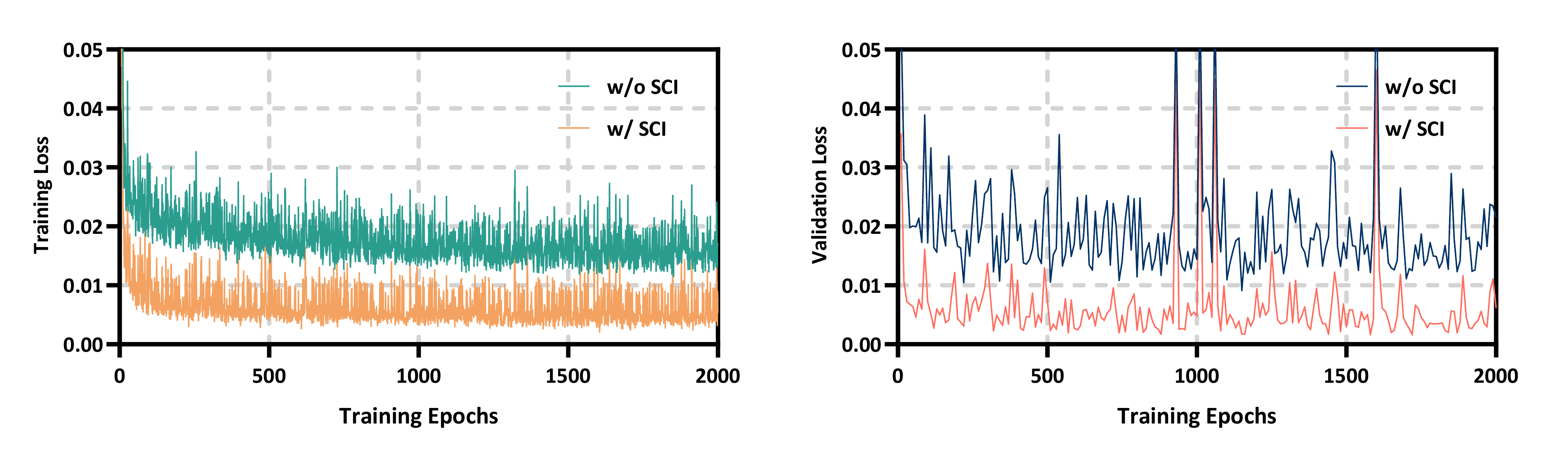}
    \caption{Loss curves over 2000 epochs. The left and right plots illustrate the training and validation losses, respectively. Incorporating SCI yields superior optimization, converging to a lower loss than the BBDM baseline (w/o SCI).}
    \label{loss}
\end{figure}

\subsubsection{Impact of the Diffusion Paradigm}

Table~\ref{table_abl} evaluates the synergistic impact of the Brownian bridge diffusion process and SCI. The standard CDM, which initiates generation from a Gaussian prior, struggles to bridge the substantial domain gap between stochastic noise and structured anatomy, resulting in suboptimal structural coherence (PSNR: 27.835 / 30.257 dB). Conversely, while the vanilla BBDM paradigm establishes a direct diffusion trajectory from the source CXR to ensure global alignment (LPIPS: 0.137 / 0.046), the absence of explicit structural guidance frequently leads to blurred pulmonary textures and degraded perceptual quality. Our DeBoneDiT model, further empowered by SCI, achieves peak performance (PSNR: 29.620 / 31.339 dB, LPIPS: 0.078 / 0.032), confirming that explicit source guidance effectively mitigates the detail loss in the bridge process while stabilizing the iterative refinement. This is corroborated by Fig.~\ref{process}, where DeBoneDiT exhibits superior retention of fine-grained anatomical details compared to both CDM and vanilla BBDM, and Fig.~\ref{loss}, where SCI facilitates more robust and accelerated convergence.

\begin{figure}[!t]
    \centering
    \includegraphics[width=\linewidth]{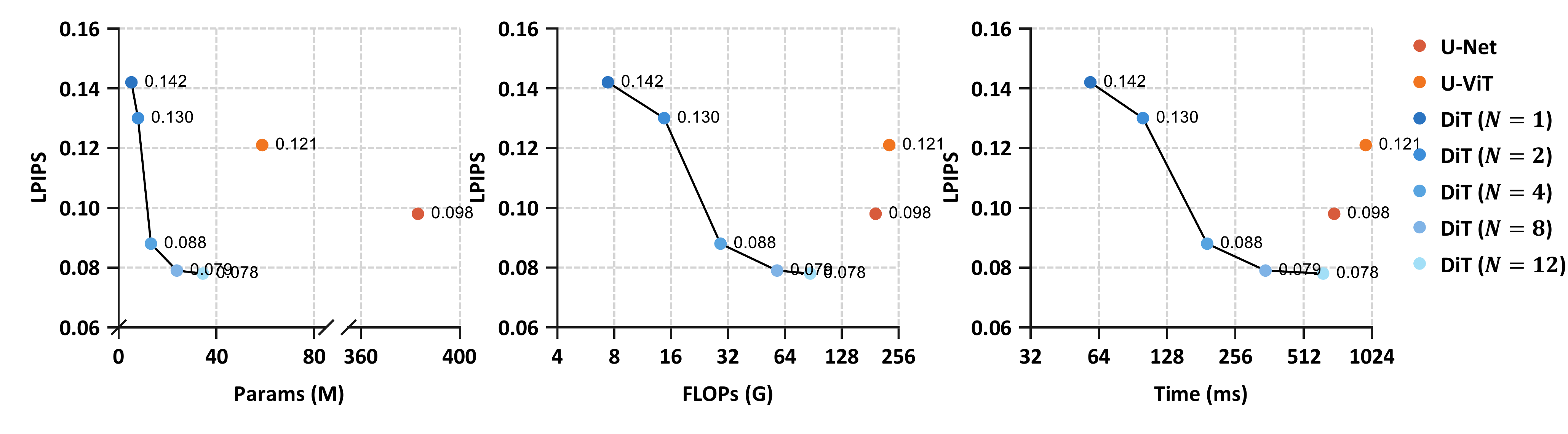}
    \caption{Quantitative ablation study on the impact of the denoising backbone. Compared to U-Net~\cite{dhariwal2021diffusion} and U-ViT~\cite{bao2023all}, the DiT-based variants~\cite{peebles2023scalable} exhibit superior efficiency and scalability while achieving state-of-the-art bone suppression performance. ($N$: number of DiT blocks)}
    \label{backbone}
\end{figure}

\begin{figure}[!t]
    \centering
    \includegraphics[width=\linewidth]{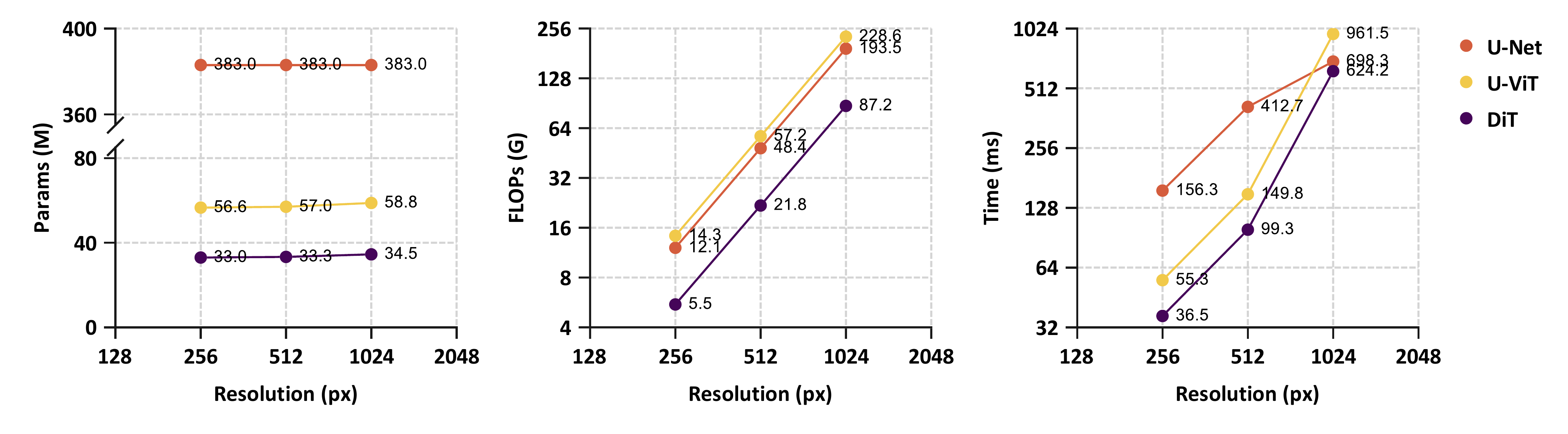}
    \caption{Quantitative efficiency comparison of various denoising backbones across different resolutions. DiT consistently outperforms U-Net and U-ViT, demonstrating robust scalability and efficiency essential for processing high-resolution medical images.}
    \label{backbone2}
\end{figure}

\begin{figure}[!t]
    \centering
    \includegraphics[width=\linewidth]{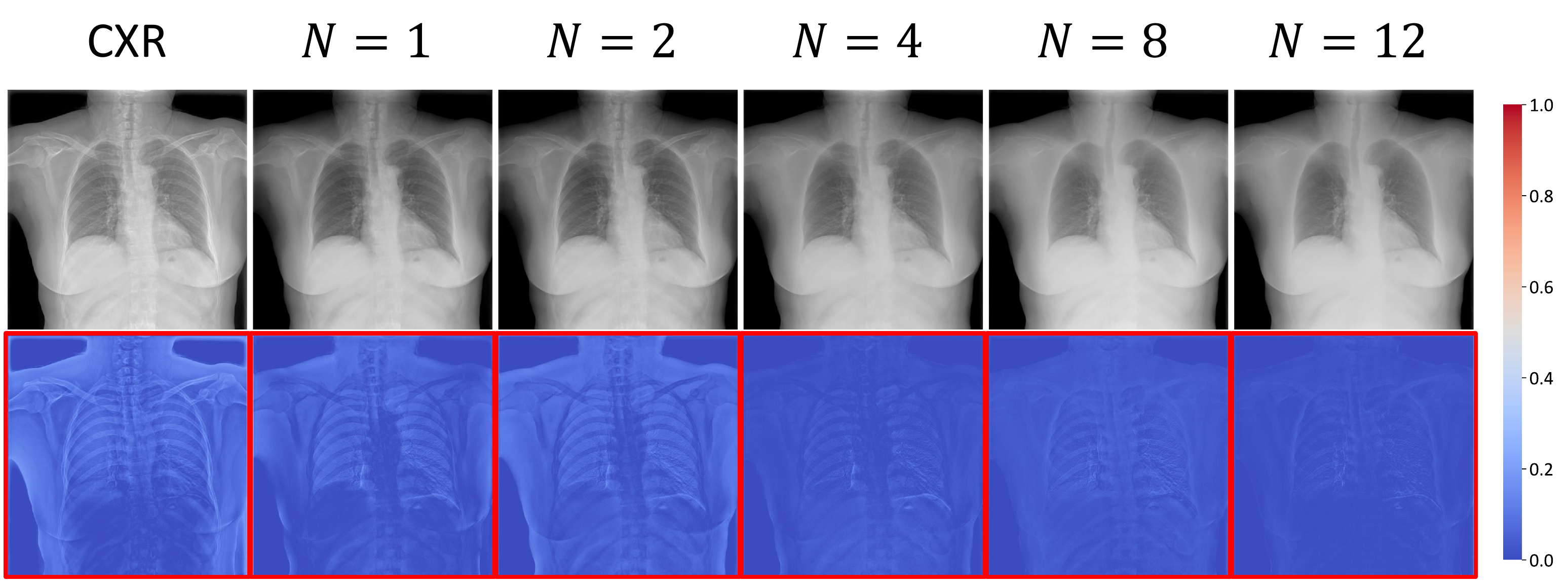}
    \caption{Visual analysis of the scaling law in bone suppression performance. The red-outlined error maps illustrate the pixel-wise residuals relative to the ground truth, which progressively diminish as the number of DiT blocks $N$ increases, indicating enhanced denoising capability.}
    \label{N_comp}
\end{figure}

\subsubsection{Impact of the Denoising Backbone}

We further investigate the influence of the denoising backbone on high-resolution bone suppression. As shown in Figs.~\ref{backbone} and~\ref{backbone2}, DiT achieves a favorable balance between generation quality and computational efficiency. Furthermore, as illustrated in Fig.~\ref{N_comp}, increasing the number of DiT blocks $N$ consistently improves the bone suppression performance, suggesting that DiT-based denoising networks exhibit strong scalability for high-resolution bone suppression.

\subsection{Downstream Evaluation}

To assess the clinical utility of DeBoneDiT, we evaluate five representative classification models~\cite{huang2017densely, szegedy2015going, radosavovic2020designing, he2016deep, zhang2018shufflenet} on the Asraf dataset. The DeBoneDiT model pretrained on the SZCH-X-Rays dataset was frozen and directly applied to generate bone-suppressed images for the Asraf dataset without further fine-tuning. Subsequently, a standard 5-fold cross-validation protocol was conducted for classification. As summarized in Table~\ref{table_down}, models trained and tested on bone-suppressed images via DeBoneDiT consistently outperform those using original CXR images across all evaluation metrics. Notably, for RegNetY-400MF~\cite{radosavovic2020designing}, the classification accuracy increases from 88.5\% to 95.5\%. Improvements are particularly evident in Recall and F1-score across different backbones, suggesting that removing overlapping bony structures facilitates the detection of subtle disease-related patterns. The consistent gains across diverse network architectures indicate that the benefits of DeBoneDiT are not tied to a specific classifier, highlighting its applicability in downstream computer-aided diagnosis.

\begin{table}[!t]
\setlength{\tabcolsep}{2pt}
\caption{Quantitative downstream classification performance on the Asraf dataset using CXRs versus bone-suppressed images, with statistical significance between paired metric values of each backbone under identical 5-fold splits evaluated using the Wilcoxon signed-rank test ($^{*}: p < 0.05$).}\label{table_down}
\centering
\resizebox{\linewidth}{!}{
\begin{tabular}{p{2.8cm}<{\centering}p{2.6cm}<{\centering}p{1.8cm}<{\centering}p{1.8cm}<{\centering}p{1.8cm}<{\centering}p{1.8cm}<{\centering}p{1.8cm}<{\centering}p{1.8cm}<{\centering}p{1.8cm}<{\centering}p{1.8cm}<{\centering}p{1.8cm}<{\centering}}
\toprule
\multirow{2.4}{*}{Model} & \multirow{2.4}{*}{Accuracy (\%, $\uparrow$)}
& \multicolumn{3}{c}{Precision ($\%, \uparrow$)} & \multicolumn{3}{c}{Recall (\%, $\uparrow$)} & \multicolumn{3}{c}{F1-Score (\%, $\uparrow$)}\\
\cmidrule{3-5}
\cmidrule{6-8}
\cmidrule{9-11}
& & Pneumonia & COVID-19 & Normal & Pneumonia & COVID-19 & Normal & Pneumonia & COVID-19 & Normal\\
\midrule
\multicolumn{11}{l}{Original CXR Images} \\ 
\midrule
DenseNet-121 & 90.8 $\pm$ 1.3 & 97.3 $\pm$ 1.2 & 96.8 $\pm$ 1.8 & 81.4 $\pm$ 2.7 & 86.5 $\pm$ 1.3 & 89.0 $\pm$ 2.5 & 96.9 $\pm$ 1.3 & 91.6 $\pm$ 0.7 & 92.7 $\pm$ 1.7 & 88.5 $\pm$ 1.8 \\
GoogLeNet & 92.3 $\pm$ 0.8 & 97.0 $\pm$ 0.8 & 96.3 $\pm$ 2.0 & 85.1 $\pm$ 1.0 & 88.7 $\pm$ 1.1 & 92.4 $\pm$ 1.9 & 95.9 $\pm$ 1.1 & 92.7 $\pm$ 0.3 & 94.3 $\pm$ 1.5 & 90.2 $\pm$ 1.0 \\
RegNetY-400MF & 88.5 $\pm$ 1.6 & 90.6 $\pm$ 4.2 & 97.7 $\pm$ 1.8 & 80.4 $\pm$ 4.3 & 90.3 $\pm$ 1.4 & 83.1 $\pm$ 5.3 & 92.0 $\pm$ 4.4 & 90.4 $\pm$ 2.3 & 89.6 $\pm$ 2.8 & 85.7 $\pm$ 2.0 \\
ResNet-50 & 91.2 $\pm$ 0.6 & 96.8 $\pm$ 3.0 & 96.2 $\pm$ 1.8 & 83.1 $\pm$ 1.6 & 86.7 $\pm$ 2.7 & 91.0 $\pm$ 2.6 & 96.0 $\pm$ 2.7 & 91.4 $\pm$ 1.1 & 93.5 $\pm$ 1.6 & 89.1 $\pm$ 0.9 \\
ShuffleNet V2 & 87.6 $\pm$ 1.5 & 90.7 $\pm$ 3.0 & 89.9 $\pm$ 5.2 & 83.1 $\pm$ 2.9 & 90.2 $\pm$ 1.4 & 89.3 $\pm$ 3.9 & 83.2 $\pm$ 8.0 & 90.4 $\pm$ 1.2 & 89.4 $\pm$ 1.5 & 82.8 $\pm$ 2.7 \\
\midrule
\multicolumn{11}{l}{Bone-Suppressed Images via DeBoneDiT} \\ 
\midrule
DenseNet-121 & 93.9 $\pm$ 0.8$^{*}$ & 98.1 $\pm$ 0.9$^{*}$ & 98.0 $\pm$ 0.6$^{*}$ & 87.1 $\pm$ 1.7$^{*}$ & 88.4 $\pm$ 1.5$^{*}$ & 94.9 $\pm$ 1.5$^{*}$ & 98.3 $\pm$ 0.7$^{*}$ & 93.0 $\pm$ 0.8$^{*}$ & 96.4 $\pm$ 1.0$^{*}$ & 92.3 $\pm$ 0.9$^{*}$ \\
GoogLeNet & 95.0 $\pm$ 0.4$^{*}$ & 97.8 $\pm$ 1.1 & 99.3 $\pm$ 0.1$^{*}$ & 89.0 $\pm$ 0.9$^{*}$ & 90.0 $\pm$ 0.6 & 96.8 $\pm$ 0.5$^{*}$ & 98.4 $\pm$ 0.6$^{*}$ & 93.7 $\pm$ 0.7$^{*}$ & 98.0 $\pm$ 0.3$^{*}$ & 93.4 $\pm$ 0.6$^{*}$ \\
RegNetY-400MF & 95.5 $\pm$ 0.3$^{*}$ & 97.8 $\pm$ 0.3$^{*}$ & 98.1 $\pm$ 1.4 & 91.0 $\pm$ 0.9$^{*}$ & 90.5 $\pm$ 0.3 & 98.8 $\pm$ 0.7$^{*}$ & 97.1 $\pm$ 1.3 & 94.0 $\pm$ 0.2$^{*}$ & 98.4 $\pm$ 0.6$^{*}$ & 93.9 $\pm$ 0.4$^{*}$ \\
ResNet-50 & 94.8 $\pm$ 0.5$^{*}$ & 98.0 $\pm$ 0.3$^{*}$ & 97.4 $\pm$ 0.8 & 89.8 $\pm$ 1.3$^{*}$ & 89.1 $\pm$ 1.3 & 97.8 $\pm$ 0.7$^{*}$ & 97.5 $\pm$ 0.6 & 93.3 $\pm$ 0.8$^{*}$ & 97.6 $\pm$ 0.6$^{*}$ & 93.5 $\pm$ 0.6$^{*}$ \\
ShuffleNet V2 & 92.5 $\pm$ 1.0$^{*}$ & 95.6 $\pm$ 3.8$^{*}$ & 97.7 $\pm$ 1.2$^{*}$ & 85.7 $\pm$ 0.9$^{*}$ & 89.2 $\pm$ 0.9 & 92.8 $\pm$ 1.2 & 95.5 $\pm$ 3.8$^{*}$ & 92.2 $\pm$ 1.4$^{*}$ & 95.2 $\pm$ 1.1$^{*}$ & 90.3 $\pm$ 1.8$^{*}$ \\
\bottomrule
\end{tabular}
}
\end{table}

\subsection{Limitations}
Despite its promising performance, DeBoneDiT remains subject to several limitations. First, owing to the inherent physical constraints of 2D projections, when pathological features are completely occluded by hyperdense bony structures, the model's reconstruction relies on learned generative priors, which may introduce hallucinations or omit extremely subtle lesions. Second, while the Depth Anything V2 encoder provides effective structural regularization, it was trained on natural images. Future work could explore CT-derived supervision to learn anatomy-aware, projection-consistent structural representations. Third, although the SZCH-X-Rays dataset establishes a high-quality benchmark, our evaluation is currently retrospective and limited to single-center data. Multi-site external validation and prospective clinical studies would further strengthen the assessment of the model's generalizability and clinical utility.
\section{Conclusion}

In this paper, we present DeBoneDiT, a depth-driven bridge diffusion Transformer achieving high-fidelity bone suppression with remarkable computational efficiency on standard CXR images. To mitigate data scarcity, we introduce SZCH-X-Rays, the largest high-quality paired dataset for bone suppression to date. By integrating structure-aware latent compression, a source-conditioned Brownian bridge diffusion process, and a DiT-based denoising network, our framework effectively preserves critical pulmonary textures while maintaining rapid inference. Extensive experiments and downstream evaluation demonstrate that DeBoneDiT achieves superior performance with reduced computational overhead, underscoring its potential as a practical tool for clinical CXR bone suppression.

\section*{Compliance with Ethical Standards}

This research was conducted retrospectively and following the principles of the Declaration of Helsinki, utilizing previously obtained images that had been anonymized before conducting analysis, thus posing no harm to patients. We applied for an informed consent waiver. Approval was granted by the Ethics Committee of Cancer Hospital \& Shenzhen Hospital, Chinese Academy of Medical Sciences and Peking Union Medical College (No. JS2023-19-2).


\bibliographystyle{plainnat}
\bibliography{main}



\end{document}